\documentclass[twocolumn]{article}

\usepackage{authblk}
\usepackage{orcidlink}
\usepackage{acronym}
\usepackage{amsfonts}
\usepackage{algorithm}
\usepackage{algorithmic}
\usepackage{graphicx}
\usepackage{rotating}
\usepackage{svg}
\usepackage{fancyhdr}
\usepackage{amsmath}
\usepackage[switch]{lineno}
\usepackage[backend=biber,style=numeric,sorting=none,natbib=true]{biblatex}
\usepackage[locale=UK]{siunitx}
\usepackage{printlen}
\uselengthunit{cm}
\DeclareSIUnit\bar{bar}

\bibliography{references}
\begin{document}

% Title
\title{Production-Ready Automated ECU Calibration using Residual Reinforcement Learning}

% Authors
\author[1]{Andreas Kampmeier\orcidlink{0009-0000-9192-6746}}%\email{kampmeier@mmp.rwth-aachen.de}
\author[1]{Kevin Badalian\orcidlink{0000-0002-5593-0227}}%\email{badalian\_k@mmp.rwth-aachen.de}
\author[1]{Lucas Koch\orcidlink{0000-0002-7368-8833}}%\email{koch\_luc@mmp.rwth-aachen.de}
\author[1]{Sung-Yong Lee\orcidlink{0000-0002-9246-8895}}%\email{lee\_sun@mmp.rwth-aachen.de}
\author[1]{Jakob Andert\orcidlink{0000-0002-6754-1907}\thanks{Corresponding Author}}%\email{andert@mmp.rwth-aachen.de}

\affil[1]{Chair of Mechatronics in Mobile Propulsion, RWTH Aachen University, Forckenbeckstraße 4, Aachen, 52074, Germany}

\date{February 26th, 2026}

\twocolumn[
    \begin{@twocolumnfalse}
        \maketitle
        \begin{abstract}
            Electronic Control Units (ECUs) have played a pivotal role in transforming motorcars of yore into the modern vehicles we see on our roads today. They actively regulate the actuation of individual components and thus determine the characteristics of the whole system. In this, the behavior of the control functions heavily depends on their calibration parameters which engineers traditionally design by hand. This is taking place in an environment of rising customer expectations and steadily shorter product development cycles. At the same time, legislative requirements are increasing while emission standards are getting stricter. Considering the number of vehicle variants on top of all that, the conventional method is losing its practical and financial viability. Prior work has already demonstrated that optimal control functions can be automatically developed with reinforcement learning (RL); since the resulting functions are represented by artificial neural networks, they lack explainability, a circumstance which renders them challenging to employ in production vehicles. In this article, we present an explainable approach to automating the calibration process using residual RL which follows established automotive development principles. Its applicability is demonstrated by means of a map-based air path controller in a series control unit using a hardware-in-the-loop (HiL) platform. Starting with a sub-optimal map, the proposed methodology quickly converges to a calibration which closely resembles the reference in the series ECU. The results prove that the approach is suitable for the industry where it leads to better calibrations in significantly less time and requires virtually no human intervention
        \end{abstract}
    \end{@twocolumnfalse}
]
\clearpage

\acrodef{MMP}{Teaching and Research Area Mechatronics in Mobile Propulsion}

\acrodef{AI}{artificial intelligence}
\acrodef{DDPG}{Deep Deterministic Policy Gradient}
\acrodef{ECU}{electronic control unit}
\acrodef{EGR}{exhaust gas recirculation}
\acrodef{ML}{machine learning}
\acrodef{MiL}{Model-in-the-Loop}
\acrodef{SiL}{Software-in-the-Loop}
\acrodef{HiL}{Hardware-in-the-Loop}
\acrodef{ViL}{Vehicle-in-the-Loop}
\acrodef{LExCI}{Learning and Experiencing Cycle Interface}
\acrodef{XiL}{X-in-the-Loop}
\acrodef{PPO}{Proximal Policy Optimization}
\acrodef{DDPG}{Deep Deterministic Policy Gradient}
\acrodef{TD3}{Twin Delayed DDPG}
\acrodef{RL}{reinforcement learning}
\acrodef{RRL}{residual reinforcement learning}
\acrodef{MDP}{Markov decision process}
\acrodef{NN}{neural network}
\acrodef{TL}{transfer learning}
\acrodef{DUT}{device under test}
\acrodef{WLTC}{Worldwide Harmonized Light Vehicles Test Cycle}
\acrodef{LP}{low-pressure}
\acrodef{HP}{high-pressure}
\acrodef{VGT}{variable-geometry turbocharger}
\acrodef{ICE}{internal combustion engine}
\acrodef{EV}{electric vehicle}
\acrodef{CCP}{CAN Calibration Protocol}
\acrodef{SBC}{single-board computer}
\acrodef{OEM}{original equipment manufacturer}
\acrodef{PID}{proportional–integral–derivative}
\acrodef{PSO}{Particle Swarm Optimization}
\acrodef{MVEM}{Mean Value Engine Model}
\acrodef{EATS}{exhaust aftertreatment system}
\acrodef{DOC}{diesel oxidation catalyst}
\acrodef{DPF}{diesel particulate filter}
\acrodef{SCR}{selective catalytic reduction}
\acrodef{CAN}{controller area network}
\acrodef{DoE}{Design of Experiments}

\section{Introduction}
\label{sec:introduction}
The use of electrical sensor and actuator systems has steadily increased in engine development in recent years.
Driven by many factors, such as stricter emissions regulations and fuel consumption requirements while simultaneously maintaining and even increasing vehicle and powertrain performance, numerous components of sensor/actuator systems have been improved and replaced. As a result, both the drive topologies and the sensor and actuator systems have become significantly more complex to adequately address the required control systems and their underlying control tasks. To achieve greater variability in solving these control tasks (e.g., active control), mechatronic systems are predominantly used today, meaning they interact both mechanically and electrically with the control loop. \citep{isermann2022automotive} This allows for a much more precise solution to the control task of the powertrain, thereby addressing the previously mentioned conflict between emissions, consumption, and performance, enabling compliance with stricter regulatory requirements.

To manage the underlying physical processes of an internal combustion engine, \acp{ECU} are employed that simultaneously capture these sensor signals in real time and actuate corresponding actuators. The primary task here is to control combustion, meaning that air and fuel supply must be adjusted to achieve an optimal balance between power, consumption, and emissions. At the same time, the exhaust aftertreatment system must be controlled so that residual pollutants can be consistently reduced at an optimal combustion ratio. This process is highly nonlinear and challenging to solve in real time due to a higher order of dependencies on multiple inputs and outputs. \citep{Cook2006powertrain}  This complexity increasingly complicates manual calibration of maps, within these controllers. Internal combustion engines continue to be controlled using map‑based strategies, as this approach constitutes a straightforward, robust, and highly reliable method for engine management. \citep{nishio2018, berger2012modeling, Atkinson2005MBCalibration}

Nowadays, control units are calibrated by powertrain experts, typically through an iterative handwritten or model-based calibration process consisting of a series of measurements, analyses, and optimization runs conducted until calibration parameters are found that meet the aforementioned requirements. Due to increased complexity of software and high dimensionality, the selection of development tools becomes increasingly important in order to make the calibration process more efficient and cost-effective. \citep{Atkinson2005MBCalibration, antinyan2020revealing} A suitable method in this context is \ac{RL}, which has already demonstrated its practical application potential in powertrain function development through various works. \citep{Shih2009, Hu2019, Ganesh22} The idea here is to automate the function development process and find an optimal control strategy for different sensor/actuator systems. The so-called agent can discover nearly optimal solution strategies through its self-adaptive capability of model-free RL algorithms without requiring human intervention. \citep{SuttonBarto}

\citet{picerno2023transfer, picerno2023ifac, KOCH2023105477} and \citet{Malikopoulos2009-jt} have already proven that an entire functional framework can practically be replaced by a reinforcement strategy. However, these works leave open how the learned knowledge could be transferred into production-ready controller calibration. The main problem remains that \acp{NN} are only partially interpretable, making it challenging to trace strategies during operation. \citep{Dulac-Arnold2021} Additionally, unrestricted interaction by an \ac{RL} agent may pose safety concerns under certain circumstances. \citep{bedei2026safeRL} Consequently, it follows that for field operation, map-based control strategies will continue to remain relevant.
\citet{LAFLAMME2025110135} propose a post-hoc explainability framework for \ac{RL}-based vehicle powertrain control, in which look-up tables are derived from a combined \ac{RL} and \ac{ML} training procedure, thereby enabling the integration of learned control policies into conventional calibration workflows.
This work proposes a systematic approach for deriving look-up tables from trained \ac{RL} agents, which satisfies safety constraints inherent to powertrain control and facilitates an automated calibration procedure that can be subsequently validated through dedicated validation runs.
This results in an iterative, but automated process, which leads to overall improvements in initial calibration while resolving the conflict among emissions, fuel consumption, and performance demands.
The paper is structured as follows: First, we introduce the fundamentals of the methodology (Sec. 2), followed by a detailed description of the design of the automated calibration pipeline (Sec.  3). Next, a concrete use case is presented to which the developed methodology is applied in a \ac{HiL} environment (Sec. 4). Subsequently, the performance of the methodology is validated under real conditions using an actual control unit (Sec. 5). Finally, the presented work is summarized, the key findings are highlighted, and an outlook on future developments is provided (Sec. 6).

\section{Methodology}
\label{sec:methodology}
In this section, the scientific methods as well as the software/hardware tools required for the
automated ECU calibration pipeline are introduced.

\subsection{Reinforcement Learning}
\label{sec:reinforcement-learning}

\Ac{RL} is a \ac{ML} paradigm concerned with training agents by allowing them to freely interact with
their environment. Based on the experiences that are generated in the process, the agent's policy is
updated to maximize the reward it receives for its behavior. Thus, \ac{RL} creates its own 
training data and --- in contrast to other paradigms --- it does not rely on pre-existing
(labeled) datasets. \citep{SuttonBarto}

The environment is modeled as a time-discrete \ac{MDP} $(S, A, P, R)$, i.e.
\begin{itemize}
    \item a state space $S$,
    \item an action space $A$,
    \item a transition probability function $P: S \times A \times S \to [0, 1]$,
    \item and a reward function $R: S \times A \times S \to \mathbb{R}$.
\end{itemize}
The agent is embodied by its policy $\pi_{\theta}: S \to A$ whose parameters $\theta$ are adjustable.
It maps observations to distributions (e.g. to Gaussians) from which actions are sampled; the policy
therefore decides the behavior of the agent. Typically, one utilizes a \ac{NN} to represent
$\pi_{\theta}$. \citep{SuttonBarto}

An interaction at timestep $t$ is fully defined by
\begin{itemize}
    \item the current state $s_{t} \in S$ of the environment,
    \item the agent's chosen action $a_{t} \sim \pi_{\theta}({} \cdot | s_{t}), a_{t} \in A$,
    \item the next state $s_{t}' = s_{t + 1} \in S$ of the environment,
    \item the reward $r_{t} = R(s_{t}, a_{t}, s_{t}')$ associated with the transition,
    \item and a flag showing whether $s_{t}$ is a terminal state ($d_{t} = 1$) or not ($d_{t} = 0$)
\end{itemize}
and stored as an experience $\chi_{t} = (s_{t}, a_{t}, s_{t}', r_{t}, d_{t})$. A series
$\tau = (\chi_{0}, \dots, \chi_{T})$ starting in an initial state and ending in a terminal one is
called an episode.  When generating training data, it is important not to always choose the action 
suggested by the policy (\emph{exploitation}), but to also stray from it on occasion 
(\emph{exploration}) in hopes of finding new avenues for solving the problem. Many libraries add 
noise (for instance sampled from a Gaussian distribution) to the output of the policy to implement
this feature. \citep{SuttonBarto}

The sum of all the rewards in an episode is defined as the return $R(\tau)$
(Eq. \ref{eq:episode-return}); it is often discounted with a factor $\gamma$. Over the course of many
training iterations, \ac{RL} algorithms attempt to maximize the expected return $J(\pi_{\theta})$
(Eq. \ref{eq:expected-return}) by slowly optimizing the policy's parameters $\theta$ via gradient ascent (Eq. \ref{eq:gradient-ascent}) or some other suitable method. \citep{SuttonBarto}
\begin{equation}
    \label{eq:episode-return}
    R(\tau) = \sum_{t = 0}^{T} {\gamma^{t} r_{t}}, \quad \gamma \in (0, 1]
\end{equation}
\begin{equation}
    \label{eq:expected-return}
    J(\pi_{\theta}) = \mathbb{E}_{\tau \sim \pi_{\theta}}[R(\tau)]
\end{equation}
\begin{equation}
    \label{eq:gradient-ascent}
    \theta_{i + 1} = \theta_{i} + \eta \nabla_{\theta}{J(\pi_{\theta_{i}})}, \quad
    \eta \in \mathbb{R}
\end{equation}

In \emph{residual} \ac{RL}, the problem to solve is tackled by combining a classical
handwritten controller $\pi_{H}$ (e.g. a look-up table/map, a rule-based controller, or a
\ac{PID} controller) and a \ac{RL} agent $\pi_{\theta}$. As a result, the overall action 
equates to the sum of the classical controller's output and the one of the RL agent:
\begin{equation}
    \label{eq:residual-rl-action}
    a_{t}(s_{t}) = \pi_{H}(s_{t}) + \pi_{\theta}(s_{t})
\end{equation}
Their contributions are not equal, though, since the nature of the problem is considered 
to be such that it can be solved using a traditional control approach for the most part; 
the agent is merely responsible for finetuning the output of the former. Because of this 
difference in magnitude, we refer to $\pi_{\theta}(s_{t})$ as a \emph{delta 
action} or simply \emph{delta} as it constitutes a small difference that is added on top 
of the main action. Residual \ac{RL} is more efficient than opting for a purely agent-based
solution; at the same time, it allows for a safer and more explainable training process as
exploration always takes place in the vicinity of the already proven controller
$\pi_{H}$. \citep{JohanninkResidualRl}

\subsection{X-in-the-Loop Platforms}
\label{sec:x-in-the-loop-platforms}
\Ac{XiL} refers to closed-loop development and testing platforms of varying virtualization levels that are employed to create/calibrate \ac{ECU} functions:
\begin{itemize}
    \item \textbf{Model-in-the-Loop (MiL)} platforms are fully virtualized; both the controller and the
    model are simulated
    \item \textbf{Hardware-in-the-Loop (HiL)} platforms consist of a physical \ac{ECU} as well as
    some select hardware components and a plant model.
    \item \textbf{Vehicle-in-the-Loop (ViL)} platforms allow a complete vehicle to interact with the
    plant model.
\end{itemize}
Manufacturers use \ac{XiL} systems as a well-established and cheap avenue for writing \ac{ECU} 
software. \citep{Kotter2018,Picerno2021,Pasternak2024,isermann2022automotive}.

\subsection{LExCI}
\label{sec:lexci}

\begin{figure}[h!]
    \centering
    \includegraphics[width=0.4\textwidth]{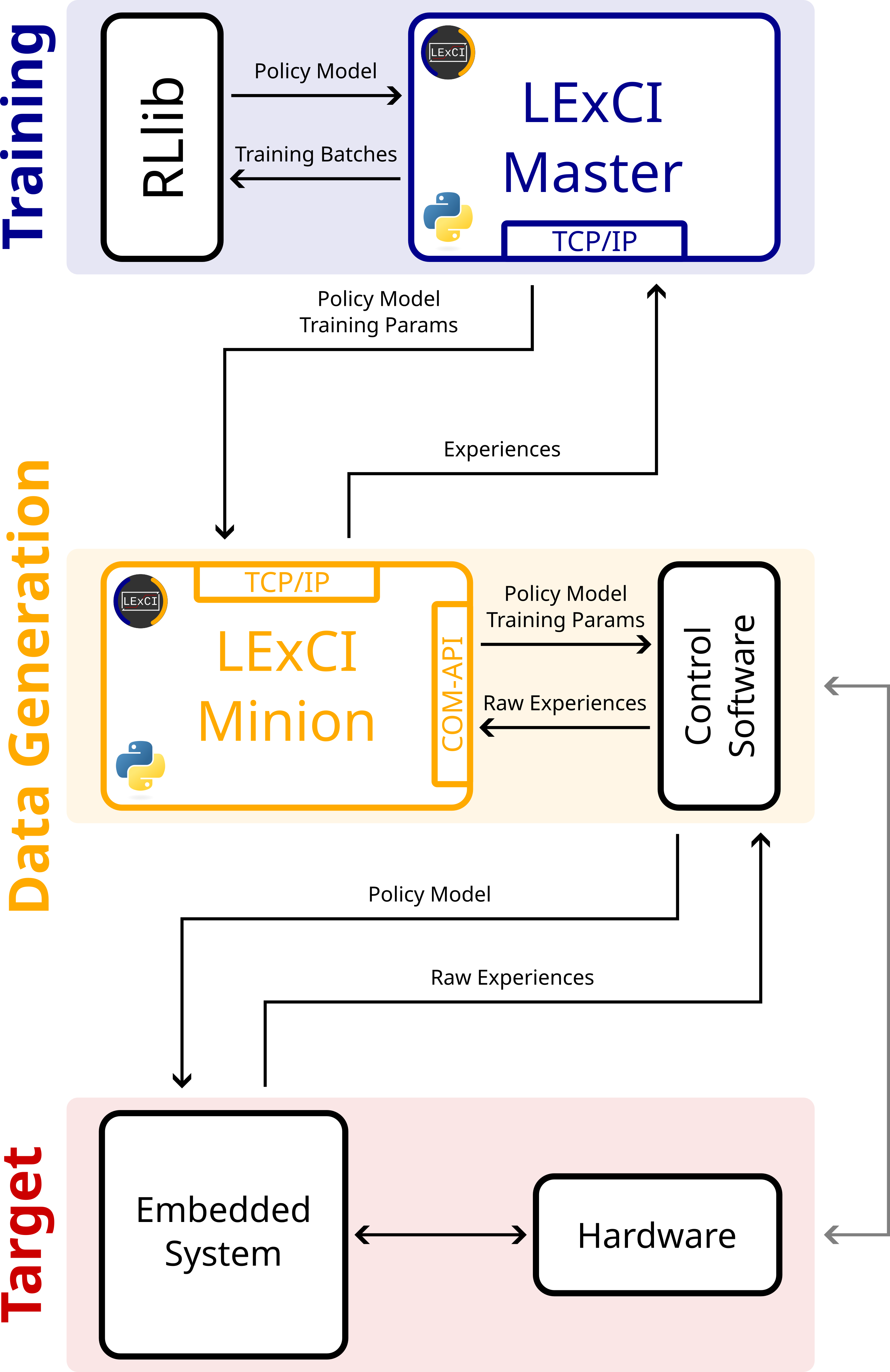}
    \caption{\Ac{LExCI}'s general software architecture (adapted from \citep{badalian2025methodology}). The communication indicated by the gray arrow is optional.}
    \label{fig:lexci-architecture}
\end{figure}

The \ac{LExCI} \citep{Badalian2024} is a framework for performing \ac{RL} with embedded systems. It 
makes use of the open-library Ray/RLlib \citep{moritz2018ray, liang2018rllib} and deploys its 
agents --- which are modeled using TensorFlow/TensorFlow Lite Micro \citep{tensorflow2015-whitepaper,
DBLP:journals/corr/abs-2010-08678} --- to embedded devices where they are executed. \ac{LExCI} 
features a master-minion architecture that separates the learning domain from the data generation 
domain (cf. Fig. \ref{fig:lexci-architecture}).

The framework comes with automation interfaces for various pieces of control/calibration software, 
including
ControlDesk\footnote{\url{https://www.dspace.com/en/pub/home/products/sw/experimentandvisualization/controldesk.cfm}},
ecu.test\footnote{\url{https://www.tracetronic.com/products/ecu-test/}}, and
MATLAB\footnote{\url{https://uk.mathworks.com/products/matlab.html}}/Simulink\footnote{\url{https://uk.mathworks.com/products/simulink.html}}. Recently, \ac{LExCI} has been extended with a 
\ac{CCP}\footnote{\url{https://www.asam.net/standards/detail/mcd-1-ccp/}} interface which grants direct access to \acp{ECU}.

\begin{figure}[h!]
    \centering
    \includegraphics[width=0.4\textwidth]{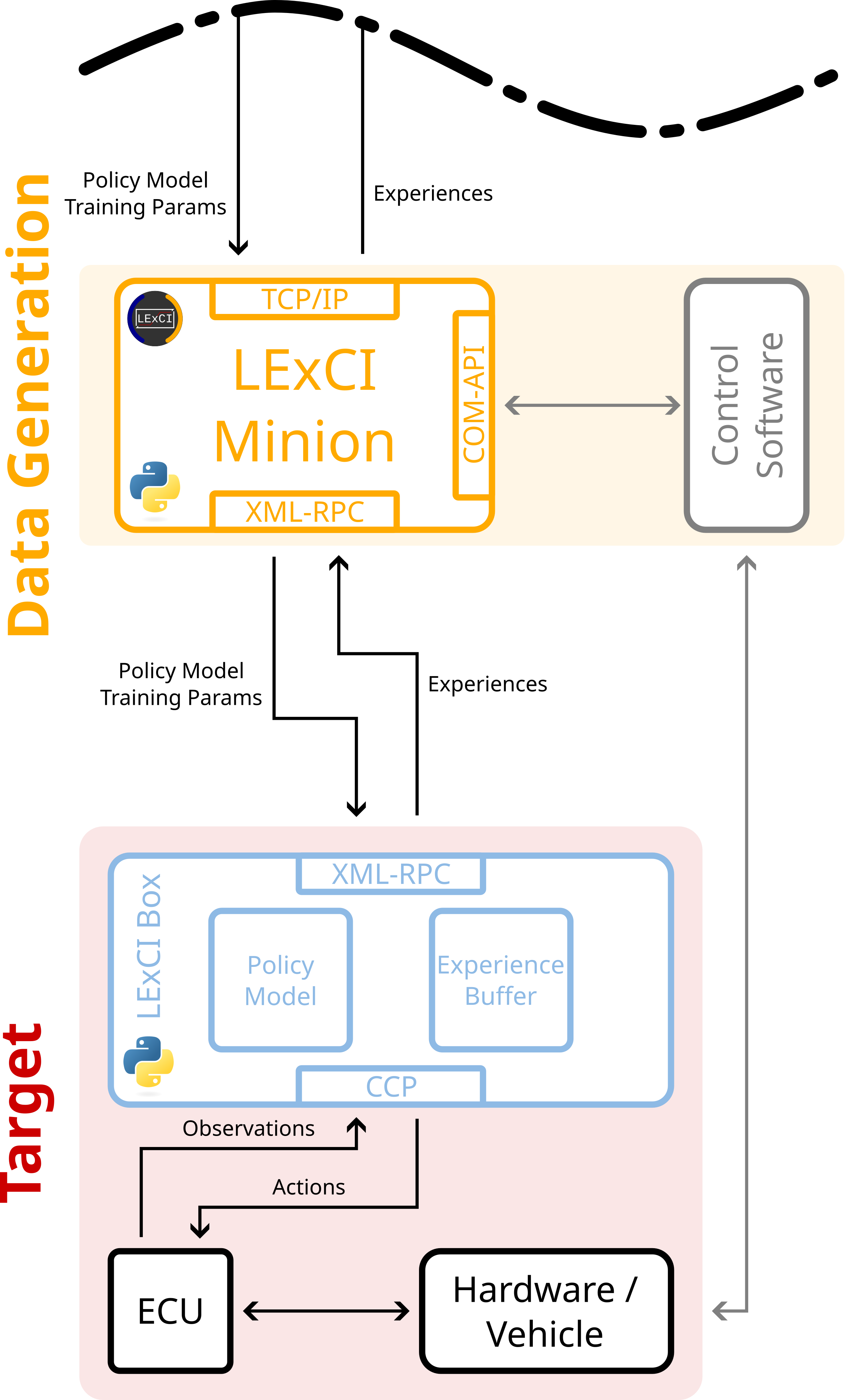}
    \caption{\Ac{LExCI}'s architecture when used in combination with the LExCI Box. The interface to the Master is not shown here as it is identical to Fig. \ref{fig:lexci-architecture}. The gray parts are optional when performing \ac{RL} in a real vehicle, but needed to configure a \ac{HiL}, for example.}
    \label{fig:lexci-box-architecture}
\end{figure}

Inspired by the software architecture of \citep{badalian2025methodology} and the insights gained in
that paper, the so called \emph{\ac{LExCI} Box} was developed: a small, standalone \ac{SBC} that can be thought 
of as an add-on to the \ac{ECU} in order to make it \ac{RL}-ready. As depicted in Fig. \ref{fig:lexci-box-architecture}, the Box connects to 
the \ac{ECU} via the aforementioned \ac{CCP} module on one side and exposes a generic interface for LExCI 
Minions to access on the other. Thus, the agent that the Box hosts can read observations directly from the 
\ac{ECU} and inject its actions into the vehicle's controller. All experiences are stored in the LExCI Box and 
can be retrieved through the Minion-side interface. Beyond that, the interface allows Minions to update the 
\ac{RL} agent on the Box, activate/deactivate stochastic sampling, and other useful helper functions.
The motivation behind the LExCI Box was twofold: First, experience has shown that, in order to learn/calibrate 
control functions properly, the agent must be able to be executed fast enough in order to keep up with the 
function in question. Although it might stand to reason to try to execute the agent \emph{on} the \ac{ECU} 
itself, that option is only available to \acp{OEM} as they generally do not grant unrestricted access to their 
hardware/software to third parties. The next best solution is the one implemented by the LExCI Box. Compared 
with the architecture presented in \citep{badalian2025methodology}, removing the calibration software and
communicating directly with the \ac{ECU} reduces latencies by up to $95\%$. Second, as explained in 
\citep{badalian2025methodology}, performing \ac{RL} in real vehicles necessitates the controller to run at all 
times. The LExCI Box fulfills this requirement.

\section{The Automated ECU Calibration Pipeline}
\label{sec:the-automated-function-development-pipeline}

\begin{figure}[!htb]
\centering
    \includegraphics[width=0.49\textwidth,keepaspectratio]{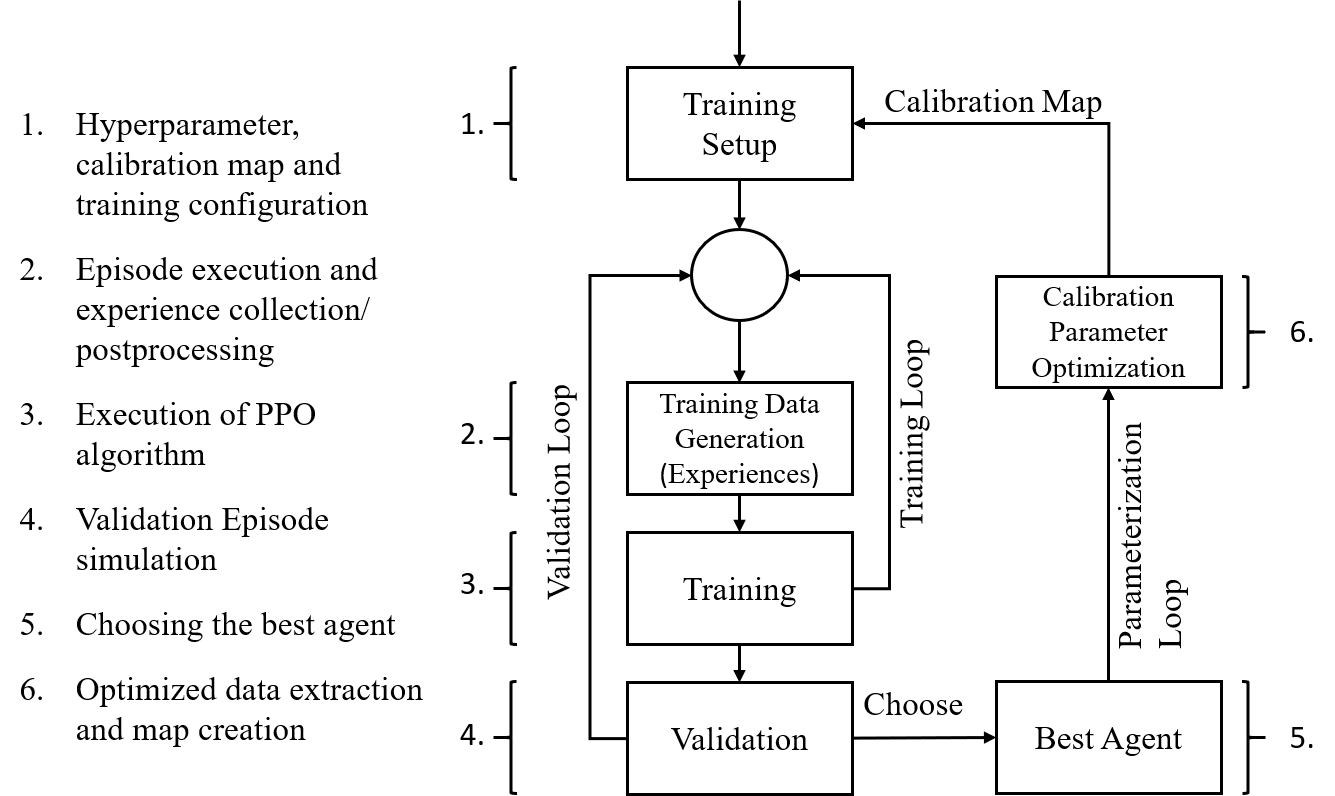}
    \caption{Flowchart of the automated ECU calibration pipeline.}
    \label{fig:automated-calibration-pipeline}
\end{figure}

As illustrated in Fig. \ref{fig:automated-calibration-pipeline}, the proposed automated \ac{ECU} calibration pipeline consists of six steps.

In the \textbf{Training Setup} phase, the \ac{ECU} function to calibrate is first examined in
several respects. One must identify the type of the control function (i.e., whether it is a map or
a classical controller like a \ac{PID}) and locate its input as well as output signals. Most 
importantly though, one has to find a way to inject the agent's action into the controller: Considering 
that full access to the \ac{ECU} is the exception rather than the rule, one needs to consult its 
documentation for unused correction maps that can be alienated in order to add the delta of the 
\ac{RL} agent to the original controller action. Once the above has been established, the \ac{RL}
algorithm's hyperparameters must be set and tuned for the specific problem; likewise, a reward function must be formulated which reflects the optimization goals. Finally, the driving 
cycles must be selected such that the agent experiences all operating points of the control 
function.

The \textbf{Training Data Generation (Experiences)} and \textbf{Training} step are responsible for 
training the residual \ac{RL} agent. They implement \ac{LExCI}'s workflow \citep{Badalian2024} to 
that end. Every batch of ten iterations, a \textbf{Validation} is performed where the agent does not 
sample its actions stochastically from the action distribution of its policy, but takes its mean 
instead. The mean is assumed to be the action that the agent would take in a certain situation; 
consequently, the validation data is more comparable because it lacks the element of chance.

Based on the validation results, the best agent is chosen in the \textbf{Best Agent} step to serve as a basis for the \textbf{Calibration Parameter Optimization} step. For map-based controllers, the policy of the agent is evaluated at the support points of the original map and the deltas then permanently added to the calibration; when applying the approach to classical controllers, their parameters must be modified by an optimization method like \ac{PSO}\citep{ParticleSwarmOptimization} such that the controller's new behavior incorporates the agent's deltas as best as possible. Both approaches will inevitably lead to a slight loss in performance compared to leaving the agent in: For maps, the policy is basically quantized which results in a loss of precision; classical controllers, on the other hand, may not be able to fully replicate the behavior of the agent. In either case, the optimized calibration makes up for it by being explainable.

The whole process can be repeated numerous times, if needed, to obtain good results. When performing residual \ac{RL}, the deltas of the agent are kept relatively small. Although this ensures that the calibration is enhanced conservatively, it also means that multiple repetitions might be necessary to optimize areas where the original controller must be altered a lot.

\section{RL Implementation for air mass setpoint calibration}
\label{sec:rl-implementation-for-air-mass-setpoint-calibration}

To demonstrate the applicability of the proposed pipeline to real-world problems, air mass setpoint calibration and optimization is selected as a use case.
The air mass setpoint calibration is a critical task of vehicle development and becomes necessary whenever hardware modifications are made to an engine. The target value determines how much air should enter the combustion chamber per stroke or injection cycle. Determining this value is particularly critical in diesel engine calibration, because combustion directly depends on the air‑fuel ratio and therefore has a significant influence on emissions, fuel consumption, and performance.
If the target value is set too low, incomplete combustion may occur, which promotes soot formation. At the same time, it can lead to power loss and increased exhaust gas temperatures. If the target value is set too high, the interaction between EGR and VTG adjustment can lead to overboosting in boost control, as the actuators operate close to their end positions. Additionally, the oxygen concentration in the intake manifold is increased, resulting in higher peak combustion chamber temperatures that promote the formation of harmful nitrogen oxides and place unnecessary thermal stress on the turbocharger.
The goal is therefore to operate the engine within the optimal lambda window, ensuring optimal and efficient combustion. This in turn supports low emission levels, which must be achieved for certification. The selection of the air mass setpoint value therefore has a particularly strong impact on the control variables of the EGR valves, the throttle valve, and the turbocharger, as it influences the amount of externally supplied air to the diesel combustion. The choice of this use case is motivated by the non-linearity and complexity of the problem and serves as a benchmark to demonstrate the potential of calibration using a residual-\ac{RL}-based strategy.
Residual \ac{RL} uses \acp{MDP} to model the interaction between the air mass setpoint value, the actuators, and the plant model that represents the agent’s environment. Defining the \ac{MDP}—including its states, actions, and reward function—requires substantial domain \mbox{knowledge} to ensure a structure that supports effective and efficient learning through meaningful data \citep{KOCH2023105477}.

\subsection{Problem Formulation for Air Mass Setpoint Calibration}
\label{sec:initial-training-in-the-model-in-the-loop-environment} 
\begin{figure}[!htb]
\centering
    \includegraphics[width=0.48\textwidth,keepaspectratio]{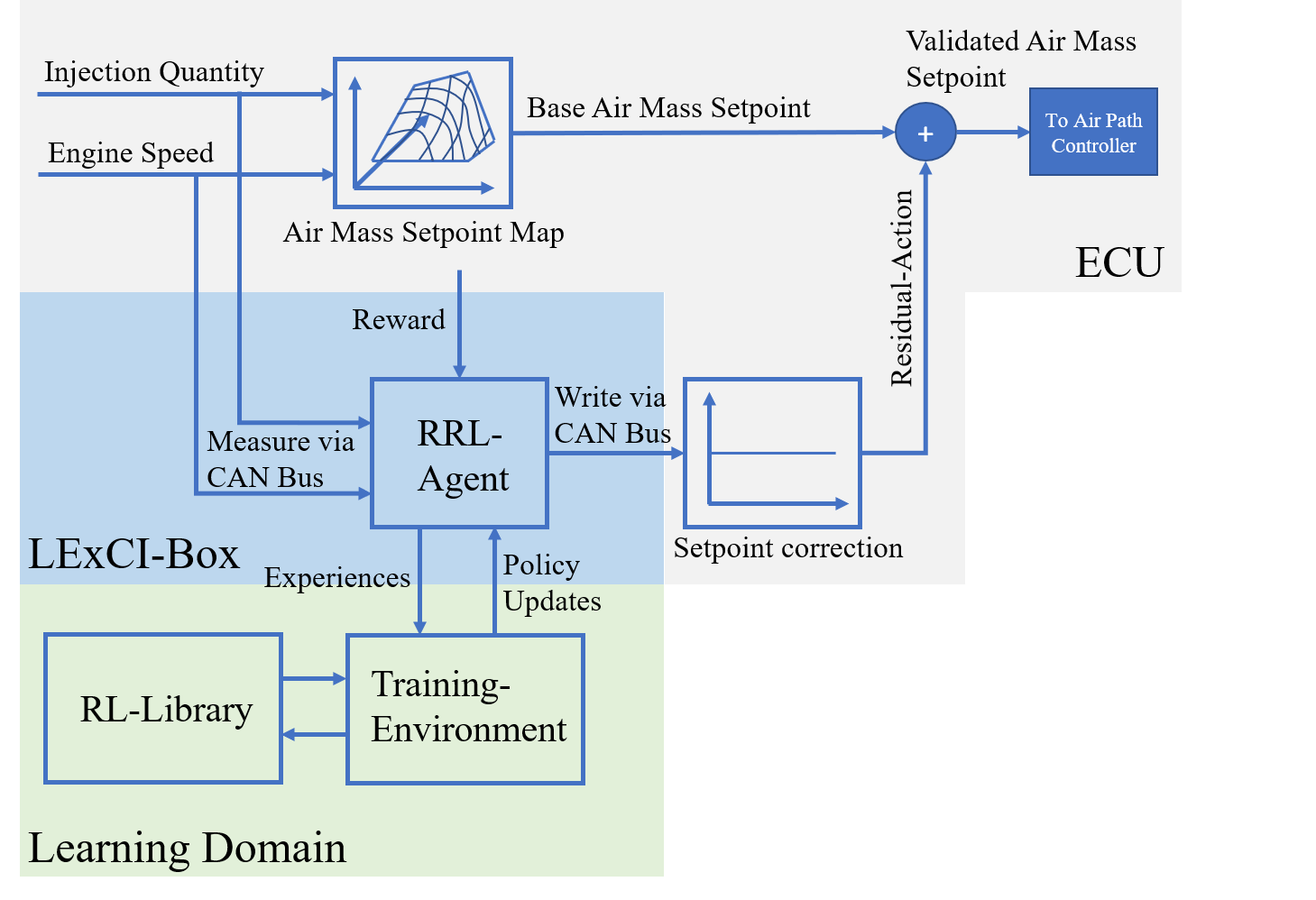}
    \caption{Air mass setpoint calibration use case}
    \label{fig:Air-Mass-Setpoint-Calibration}
\end{figure}

\Ac{RRL}, due to its self‑learning nature, offers the possibility of adaptively optimizing the target values during driving operation. This provides an advantage over the classical calibration methodology, which requires many driving tests and manual adjustments.
For this purpose, an intervention is performed in parallel with the control unit’s target‑value determination. The scheme is shown in Fig. \ref{fig:Air-Mass-Setpoint-Calibration}. The air mass setpoint determination typically consists of a base map and correction values that can be activated under given external conditions. These include, for example, temperature and air pressure changes during operation. Furthermore, dynamic corrections usually exist for acceleration maneuvers. The result is a validated air mass setpoint that is passed to the air‑path controller. The air‑path controller then adjusts the target air mass via actuator control (\ac{EGR}, throttle, and \ac{VGT}) which controls the actual air mass entering the engine.

If a \ac{RRL} agent is implemented in parallel with the air mass setpoint determination, a difference value can be provided as an action by measuring the map inputs and determining the current reward. This difference value is written back to the control unit in soft real-time via a calibration variable and added to the current setpoint value. This yields a corrected setpoint value which, through additional driving cycles, leads to new experiences that again influence the reward and thus lead to new actions. It is important that the adjustment of the setpoint value represent only a fraction of the overall value in order to ensure safe training in vehicle operation. Through this adaptive and self‑learning behavior, optimal parameters for the setpoint value can be found.
A disadvantage of delta‑based (see Sec. 2.1) adjustment is the possibility of ending up in a local minimum. Varying the action space can provide a remedy here. To conduct successful training, a reward must additionally be defined for the \ac{RRL} agent. The reward should generally be designed to include the quantities to be optimized for determining the target variable. These include, for example, boost pressure, NOx emissions, and soot emissions. These are typical optimization quantities also used in traditional calibration methods. For this reason, the implemented reward function was designed so that deviations in boost pressure and emissions are penalized. The magnitude of the penalties for each quantity is weighted using proportionality factors:
\begin{equation}
    \label{eq:xil-reward-function}
    \begin{split}
        R_{\text{XiL}}(s, a, s') = & -\alpha_{1}
        \dot{m}_{\text{NO\textsubscript{X}}} -\alpha_{2} \dot{m}_{\text{soot}} \\
        & - \alpha_{3} |\Delta\text{p\textsubscript{Boost,dev}}|
    \end{split}
\end{equation}
A stability term is omitted in the reward function because, due to the delta‑based adjustment, it is unlikely that engine stalls or unsafe states will occur. In summary, the problem to be approximated is characterized in Table \ref{tab:egr-controller-agent-io}. The inputs of the calibration map are treated as system states, whereas the feedback signals are derived through the reward function. These signals comprise the boost‑pressure deviation as well as the NOx and soot mass flow rates, all of which are determined by the engine control unit. The output of the function is a differential setpoint air mass value, which is subsequently added to the current target value within the control unit.

\subsection{Training in the Hardware-in-the-Loop Environment}
To verify the applicability of the proposed methodology under near‑real conditions, a \ac{HiL} system was selected, as the control unit and its embedded control structures are available as real components. Therefore, the \ac{HiL} provides an environment that enables the agent to accumulate meaningful experience. This makes it a highly suitable development platform, enabling an almost seamless transfer of the methodology to test benches or directly to the vehicle at a later stage. The challenge in \ac{HiL} simulation lies in the fidelity of the individual models, which must meet high accuracy requirements while simultaneously maintaining real‑time capability. Furthermore, environment, powertrain, and vehicle models are required to conduct realistic driving profiles. The following sections will introduce the powertrain models, with a focus on the engine and emissions models, as these are essential for validating the proposed methodology.
The target vehicle and its powertrain belongs to the D-segment of the European car classification
system. The engine is equipped with a single-stage \ac{VGT} and \ac{LP} as well as \ac{HP} \ac{EGR} paths. All models were calibrated using both engine test bench and vehicle roller test bench measurements from a \ac{WLTC} reference test. Further specifications are summarized in Tab.
\ref{tab:target-vehicle-specifications}.

\begin{table}[h]
\sisetup{range-phrase = --,range-units = single}
  \centering
  \begin{tabular}{| l r |}
    \hline
    Engine: & inline four-cylinder Diesel \\
    Displacement: & \SI{1999}{\centi\meter^3} \\
    Maximum power: & \SIrange{120}{130}{\kilo\watt} @ \SI{4000}{\minute^{-1}} \\
    Maximum torque: & \SIrange{380}{450}{\newton\meter} @ \SI{1750}{\minute^{-1}} \\
    Transmission: & eight-speed automatic \\
    Curb weight: & \SI{1590}{\kilogram} \\
    \hline
  \end{tabular}
  \caption{Specifications of the target vehicle.}
  \label{tab:target-vehicle-specifications}
\end{table}

To accurately capture the dynamic behavior of an internal combustion engine during transient driving maneuvers, a physics‑based \ac{MVEM} was employed. This modeling approach provides the capability to compute all relevant thermodynamic states and sensor‑level quantities required by the \ac{ECU}. The model determines, among other variables, manifold pressures, mass flow rates, and temperatures, using a volumetric‑efficiency‑based filling model that represents the dominant gas‑exchange processes at a mean‑value level.
The \ac{EGR} control paths are formulated using a map‑based methodology that predicts valve flow characteristics and associated loss curves. The prediction of NOx and soot emissions relies on a semi‑physical modeling framework that couples empirical calibration maps with physically motivated correlations \citep{querel2015semiPhys}, thereby enabling a computationally efficient yet sufficiently accurate representation of pollutant formation processes.
The turbocharger subsystem is represented through a thermodynamic power‑balance model that links compressor and turbine performance. The turbocharger shaft speed is inferred from the compressor power, after which turbine power is computed using the turbine rotational speed and exhaust backpressure. This formulation ensures a physically consistent representation of turbocharger dynamics under transient load conditions.
The overall engine model receives, as boundary conditions, the injection timing signals (start and end of injection), the full set of actuator positions, and the actual engine speed provided by the crankshaft dynamics model as well as environmental conditions. The crankshaft model itself is coupled to a virtualized automatic transmission, incorporating a torque converter and lock‑up clutch, which in turn interfaces with the drivetrain. The drivetrain then actuates a vehicle model that represents vehicle mass properties and rotational inertias, thereby reproducing realistic load conditions during acceleration and deceleration events.
A corresponding \ac{EATS} model represents the dynamic behavior of the \ac{DOC}, the \ac{DPF}, and the \ac{SCR} subsystem with urea dosing. Collectively, the \ac{MVEM}, the \ac{EATS} model, and the drivetrain models have a well‑established record of robustness and fidelity for virtual \ac{ECU} calibration workflows, as demonstrated in prior studies. \citep{Lee2018HiLforRDE}

\begin{table*} [h!] 
\centering
  \begin{tabular}{| r | c | l |}
    \hline
    \textbf{Number} & \textbf{Input} & \textbf{Description}  \\
    \hline
    1 & $n_{\text{eng},t}$ & Engine speed \\
    2 & $m_{\text{inj,tot},t}$ & Total injected quantity per stroke \\
    \hline
    \hline
    \textbf{Number} & \textbf{Output} & \textbf{Description} \\
    \hline
    1 & $\mu_{\dot{r}_{\text{m}\textsubscript{air,cor}, t}}$ & Mean of the Gaussian air mass for setpoint correction  \\
    2 & $\sigma_{\dot{r}_{\text{m}\textsubscript{air,cor}, t}}$ & Standard deviation of the air mass setpoint correction\\
    \hline
    \hline
    \textbf{Number} & \textbf{Reward} & \textbf{Description} \\
    \hline
    1 & $\dot{m}_{\text{NO\textsubscript{X}}}$ & NOx mass flow based on Eq. \ref{eq:xil-reward-function} \\
    2 & $\dot{m}_{\text{soot}}$ & Soot mass flow based on Eq. \ref{eq:xil-reward-function} \\
    3 & $|\Delta\text{p\textsubscript{Boost,dev}}|$ & Boost pressure error for governor control based on Eq. \ref{eq:xil-reward-function} \\
    \hline
  \end{tabular}
  \caption{Inputs, outputs and reward inputs of the agent's policy for setpoint calibration.}
  \label{tab:egr-controller-agent-io}
\end{table*}

\section{Performance Evaluation}

To demonstrate the applicability of the methodology presented in Chapter \ref{sec:rl-implementation-for-air-mass-setpoint-calibration} in a real environment, the previously introduced use case of air mass setpoint value calibration was selected. In doing so, the control structure of the \ac{EGR} valve control remains unchanged, i.e., the control unit’s strategy for controlling the air mass setpoint value remains active, even though an \ac{RL} agent intervenes. The results presented in this section were obtained using a \ac{PPO}\citep{DBLP:journals/corr/SchulmanWDRK17} agent that was trained on a defined segment of the \ac{WLTC}. For this work, training was carried out exclusively in a \ac{HiL} environment, since \ac{HiL} methodologies represent an optimal balance between development costs, debugging capabilities, and fidelity to the target vehicle application. Additionally, the \ac{HiL} environment is particularly safe with respect to unexpected behavior. Because a real control unit is integrated, the new methodology can be tested under highly realistic conditions, which allows demonstrating the robustness of the presented approach. It is therefore particularly suitable for the development of new concepts for calibration methodologies and simultaneously provides realistic, \ac{ML}‑generated characteristic maps that can potentially be used in vehicle operation. The evaluation of the training progress in \ac{HiL} reveals the overall performance of each trained and validated agent compared to to a reference strategy. The approach includes a multi-criteria analysis of the optimal agent based on the conflicting criteria of performance and emissions benefits.

\subsection{Training process with real control unit}

\begin{figure}
    \centering
    \includegraphics[width=0.49\textwidth,keepaspectratio]{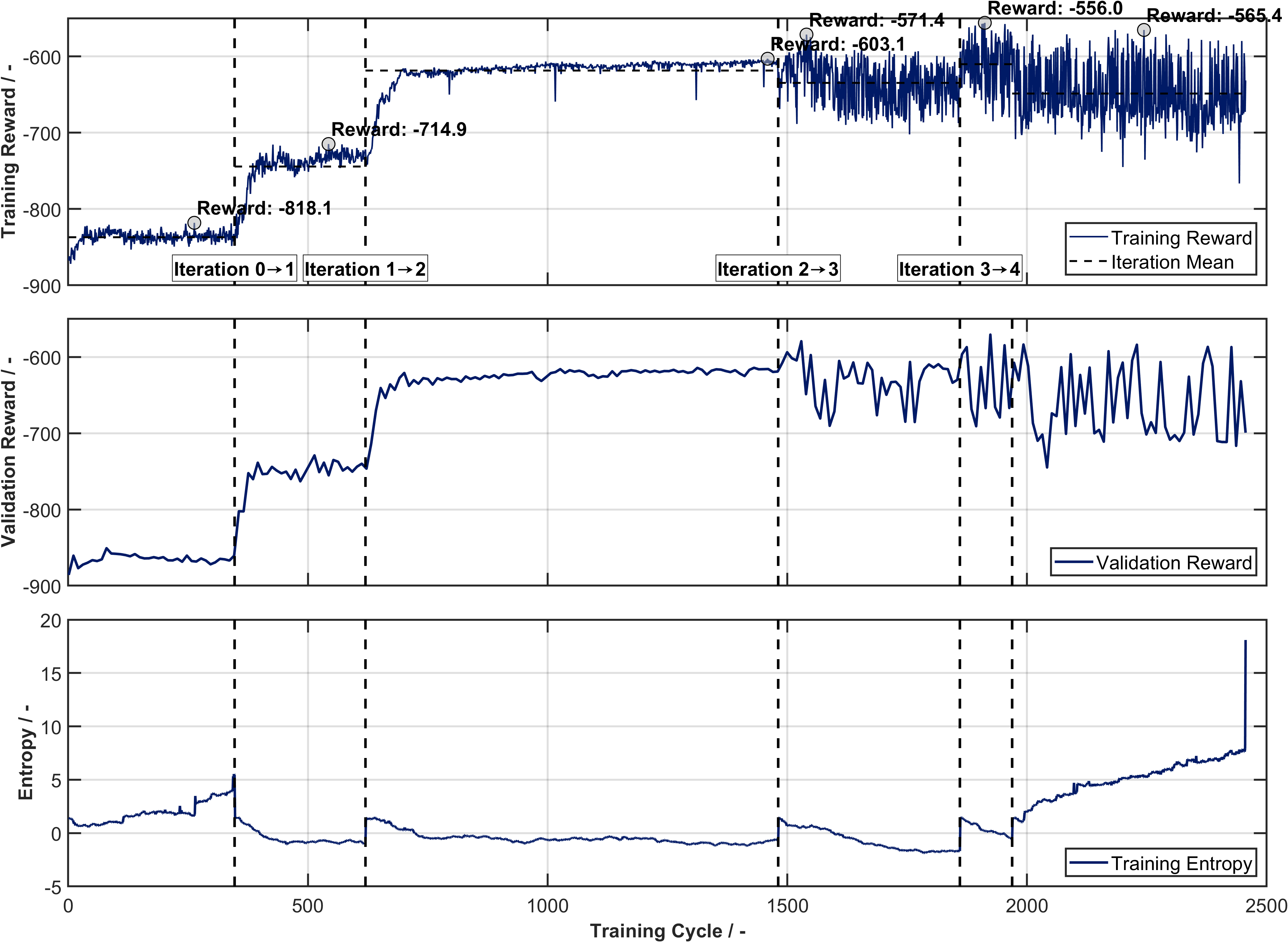}
    \caption{Learning Process - cumulative reward and entropy in HiL training.}
    \label{fig:Learning-Process}
\end{figure}

The training sessions used a 430.9 second segment of the \ac{WLTC}, during which an average speed of 37.6 km/h was achieved. This ensured that the operating points were primarily optimized in the \ac{EGR} relevant part‑load range. The training itself required a total duration of 116 hours and was conducted on an Intel Xeon E5‑1630 v4 quad‑core processor. The connected NVIDIA Jetson Orin Nano hosted the LExCI Box, i.e. it executed the corresponding policy network in real-time and interacted with the control unit via \ac{CAN} communication. The \ac{SBC} accumulated a calculated engine operating time of 295 hours, during which the agent collected experiences with the plant models and executed actions that were subsequently used for training.
A main indicator for evaluating the training process is the cumulative reward over all episodes during an entire calibration iteration. In total, five iteration runs (map-calibration sessions) were carried out, which are shown in Fig. \ref{fig:Learning-Process}.
It can be seen that at the beginning of the first iteration run, the reward initially increases steeply and a converging trend develops, stabilizing at an average value of –840, before dropping slightly again. The reward of the validation runs initially follows this trend but then begins to decrease while entropy\footnote{In this context, entropy represents a measure of the agent’s uncertainty in action selection.} is slightly increasing simultaneously. Since the agent with the highest validation reward was still in a phase of decreasing entropy, the run could still be considered valid and was therefore used for characterizing the calibration map for the next run.
The next two iteration runs show a strongly converging behavior while entropy decreases at the same time, which indicates an increasing maturity level of the agent accompanied by a decreasing degree of exploration. The average reward rises to –620 in this process. Iteration runs four and five continue to show decreasing entropy, but the reward begins to fluctuate in both the training and validation runs. This can most likely be attributed to the proximity of a local minimum, causing the agent to no longer be able to significantly increase the reward with the learned action and therefore begin exploring again.
The training was then intentionally continued until the rising entropy indicated that the agent was increasingly reverting to exploration and the uncertainty in its action selection was growing. With the presence of both indicators (uncertainty in action selection and unstable reward), it was reasonable to assume that training could be concluded at this point. The result is the best-performing agent, shown in Fig. \ref{fig:Transient-Results}. In total, 2,463 training cycles were executed, corresponding to a total training time of 411 hours in real time.
\begin{figure}
    \includegraphics[width=0.48\textwidth,keepaspectratio]{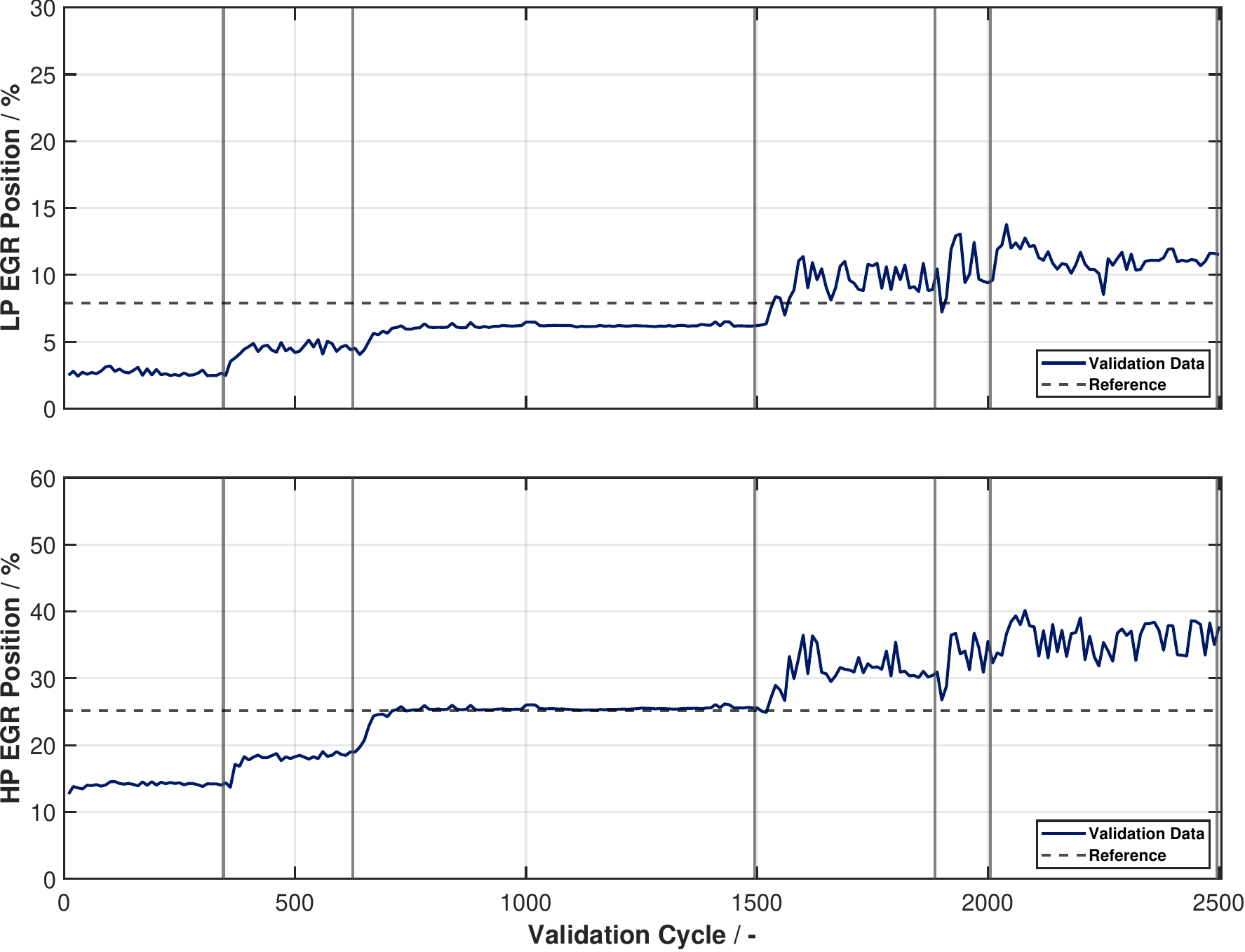}
    \caption{Learning Process - Average HP-EGR and LP-EGR position in HiL-Training.}
    \label{fig:Learning-Process-EGR}
\end{figure}

Another indicator that supports the maturity of the agent’s strategy is the progression of the average \ac{EGR} positions during the validation cycles which can be seen in Fig. \ref{fig:Learning-Process-EGR}. Starting from an initial baseline map, only small \ac{EGR} valve positions are commanded at the beginning of the training. As the training progresses, the positions are gradually increased until alignment with the reference strategy is achieved. The increase in valve positions has a direct influence on NOx and soot emissions, with NOx being the dominant driver in this case. By increasing the valve positions, more exhaust gas is recirculated, which directly lowers the peak combustion temperature and therefore results in reduced NOx production. This increases the reward, providing the agent with feedback that the chosen actions are beneficial.
At the end of the training, the valve positions are slightly higher than the reference and average just above 30 \%  for the \ac{HP} \ac{EGR} and 10 \% for the \ac{LP} \ac{EGR}. The reference strategy is at 25 \% for the \ac{HP} \ac{EGR} and 8 \% for the \ac{LP} {EGR} respectively.
\begin{figure} [ht]
    \centering
    \includegraphics[width=0.48\textwidth,keepaspectratio]{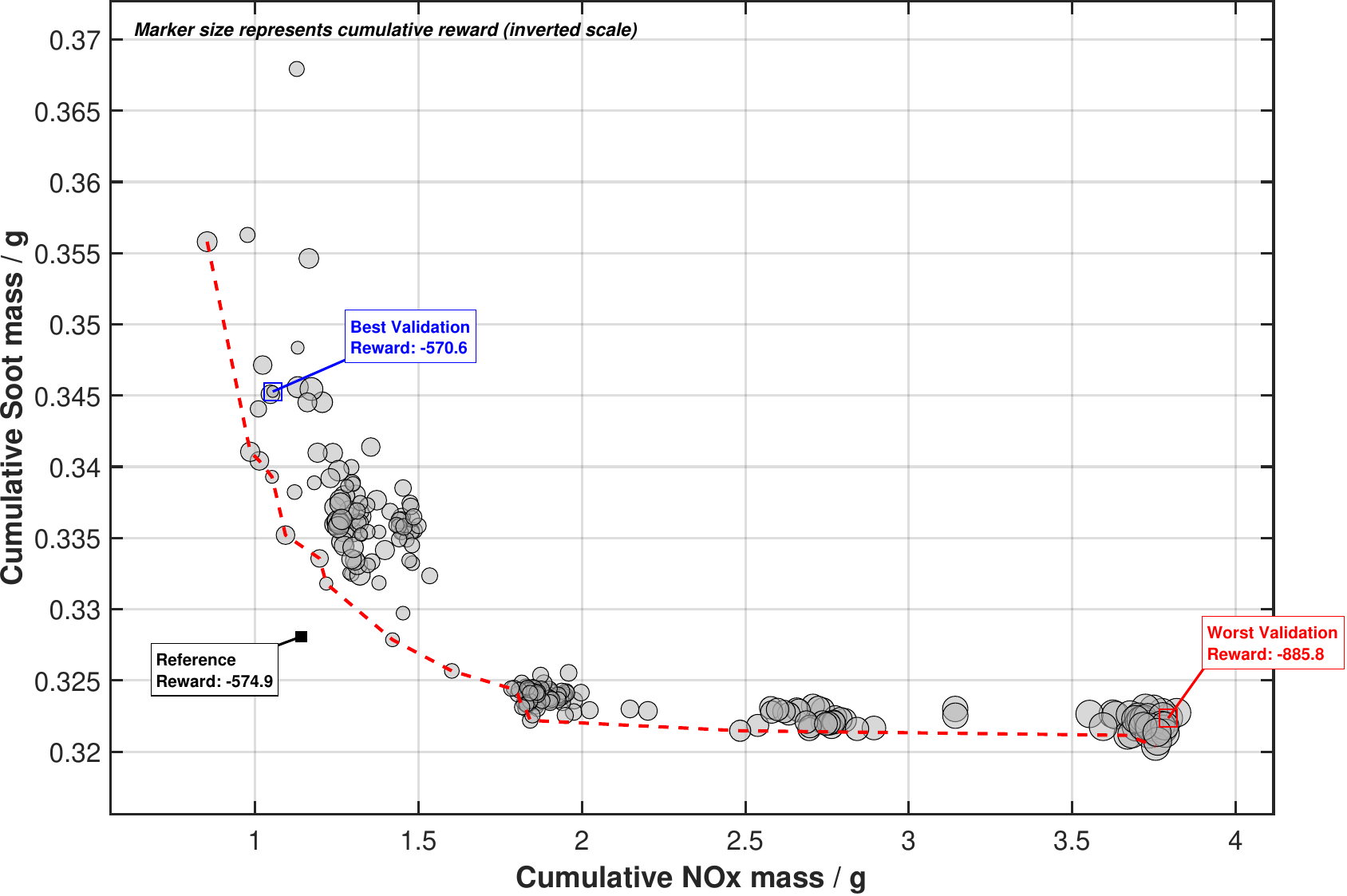}
    \caption{Validation results of WLTC section for NO$_{x}$, soot and cumulative reward in function of the performed calibration iterations.}
    \label{fig:Pareto-Plot-NOx-Soot}
\end{figure}
The values presented indicate a successful training process, but they do not yet allow an evaluation of the agent’s performance with respect to its multi-objective reward function. For this purpose, the training progress is shown in Fig. \ref{fig:Pareto-Plot-NOx-Soot}. The diagram depicts the well‑known NOx/soot trade‑off, where each point represents an validation episode. The position of the point corresponds to the magnitude of the cumulative engine‑out NOx and soot, which are captured directly from the control unit via the introduced software‑hardware interface. The inverted size of the points represents the reward achieved in that episode. 
The only slight actuation of the \ac{EGR} valves at the beginning of the training leads to a lower cumulative soot value, but to high NOx emission levels ranging between 3.5 and 4 g. As the recirculated exhaust gas mass flow increases, the NOx value decreases from iteration to iteration, but at the cost of increasing soot levels. A Pareto front forms, but it does not reach the NOx/soot values of the reference. One reason for this is that the boost pressure deviation is an additional input to the reward function to reflect the performance dependency. The reference strategy does not directly account for this dependency by a weighted reward factor and may tolerate greater boost pressure deviations. Therefore, under these boundary conditions, the agent finds the best cumulative reward only above the reference, meaning lower NOx values but simultaneously higher soot values. Finally, values of 1.05 g for NOx and 0.345 g for soot could be achieved for a total cumulative reward value of -570.6. The reference is similarly good, but the reward value is slightly smaller than the best run of the best agent with a total cumulative reward of -574.9.

\subsection{Agent Validation in Transient Condition}
\begin{figure} [t!]
    \includegraphics[width=0.49\textwidth,keepaspectratio]{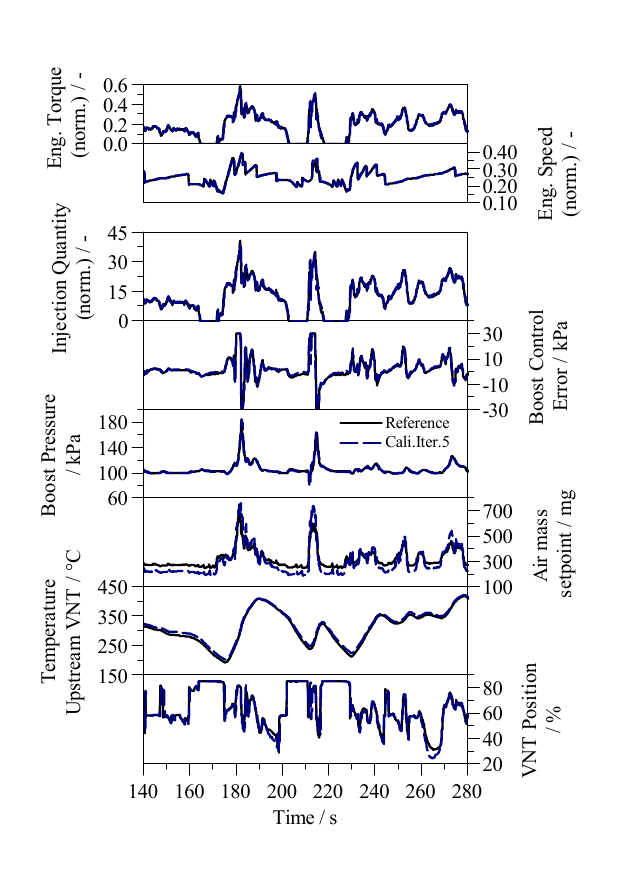}
    \caption{Validation of cycle results between calibration iteration five and reference calibration in WLTC section.}
    \label{fig:Performance-Analysis}
\end{figure}

\begin{figure*}
    \centering
    \includegraphics[width=\textwidth,height=0.88\textheight,keepaspectratio]
        {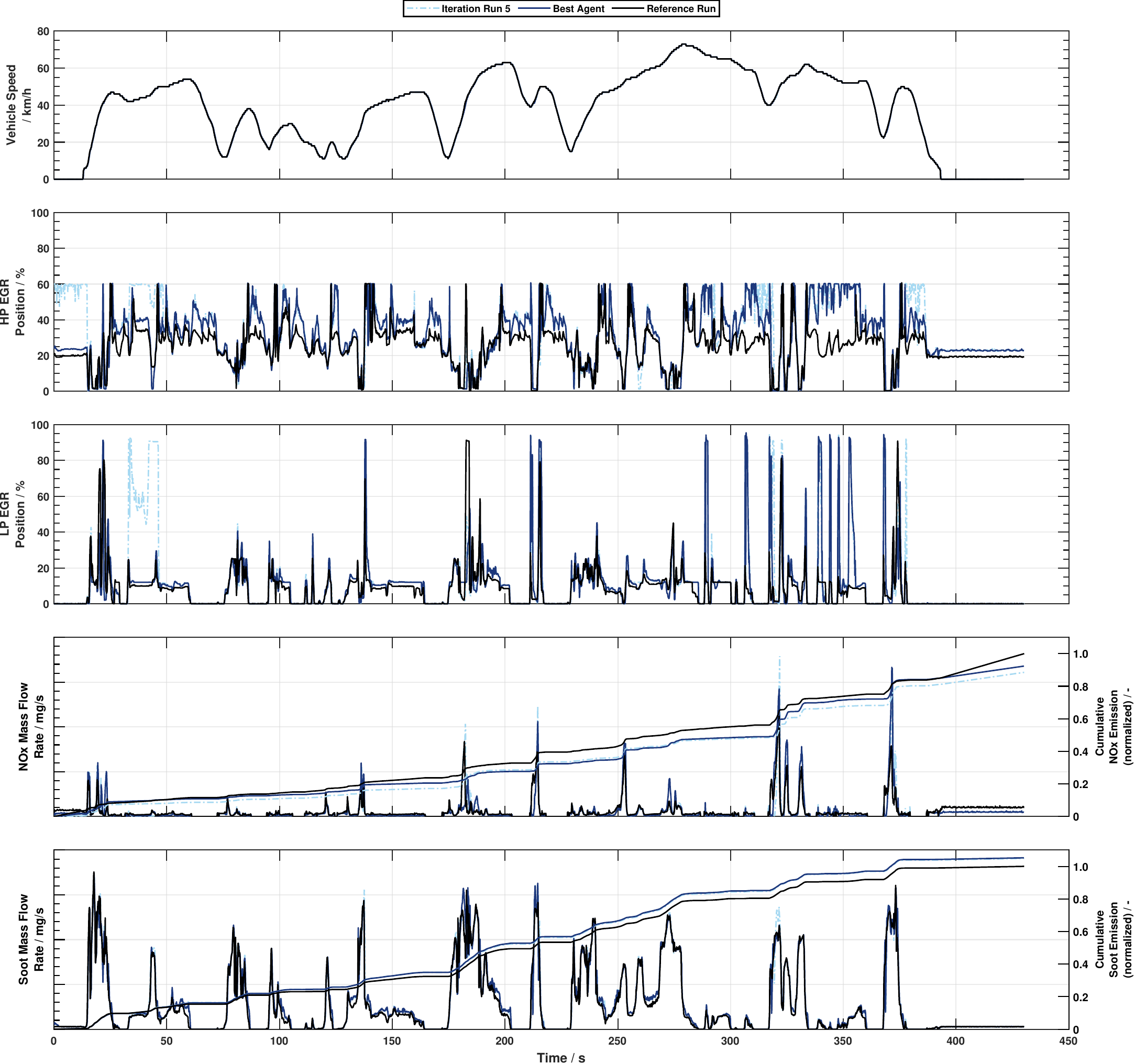}
    \caption{Comparison of EGR positions, NOx and soot mass flows, as well as cumulative values, 
    between best validation runs and reference measurement in WLTC section for different calibration stages}
    \label{fig:Transient-Results}
\end{figure*}

To assess the learned strategies not only from a global and cumulative perspective but also in a time‑resolved, application‑relevant manner, three different validation runs are presented in Fig. \ref{fig:Transient-Results}. These include a reference run using the baseline control strategy, a run employing an iteratively calibrated map, and a run using the best-performing agent that achieved the highest reward. The top plot shows that all strategies are able to follow the prescribed speed profile, indicating that none of the approaches lead to a noticeable deterioration in the overall drivability or performance capability of the vehicle. It should be noted, however, that the operating points within this cycle are predominantly characterized by low to medium load conditions.
Analyzing the actuator positions of the exhaust gas recirculation valves reveals that the newly learned air mass setpoint values result in generally increased actuator levels. A closer inspection shows that, particularly during transitions into fuel cut‑off, the agent has learned to command lower air mass setpoint values, which leads to a continuous reduction in NOx emissions. During positive torque demand, both strategies exhibit remarkably similar behavior, differing by only a few percentage points in actuator magnitude. This suggests that the iterative calibration using an \ac{RL} agent has already produced a strategy that is viable for practical application.
Overall, both the best agent and the iteratively calibrated map achieve lower cumulative NOx emissions compared to the reference. A drawback arising in conjunction with the optimized boost‑pressure control is that the learned strategy accepts slightly higher soot levels. In total, the best agent marginally surpasses the reward achieved by the reference strategy by 4.3 points. Notably, transient NOx peaks are greater than those of the reference approach, but these are compensated by reduced NOx levels across the remaining portions of the driving cycle.
To evaluate the quality of the derived calibration and its strategy, a defined section of the driven cycle was compared between the obtained calibration and the reference. In Fig. \ref{fig:Performance-Analysis} it becomes evident that, when considering performance relevant parameters such as torque and boost pressure, there is a good agreement between the \ac{RL} based calibration and the reference strategy. Additionally, the fuel consumption, represented as the injected fuel quantity per stroke, also shows an almost identical progression. Differences can be observed in the selected air mass setpoint, which results in a different \ac{EGR} mass flow and consequently leads to variations in the inducted fresh air mass. It is clearly visible that the agent pursues a different strategy for NOx reduction. In phases of very low injection quantities and overrun phases, it significantly reduces the air mass setpoint, while during acceleration phases it tends to aim for slightly higher setpoints. As can be seen in Fig. \ref{fig:Transient-Results}, these differences ultimately balance out, overall leading to lower NOx emissions compared to the reference. The \ac{VGT} positions also show good agreement with a tendency toward slightly lower positions compared to the reference. This demonstrates that the derived calibration, while not necessarily superior, can compete on an equal level with the reference calibration.

\begin{figure*}[h!]
    \centering
    \includegraphics[width=\textwidth,height=0.5\textheight,keepaspectratio]
        {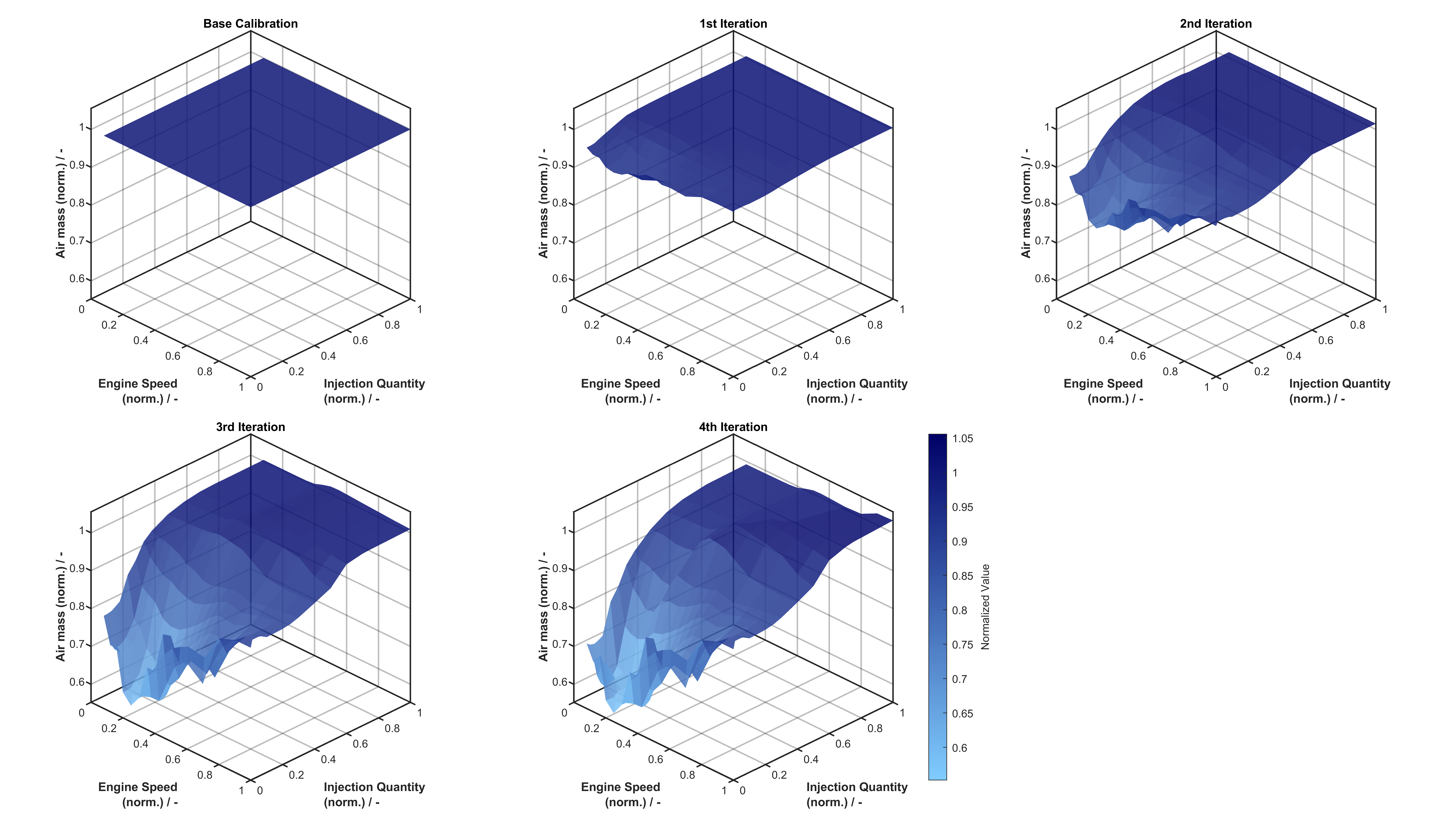}
    \caption{Calibration progress after four calibration iterations compared to base calibration}
    \label{fig:Calibration-Progress}
\end{figure*}
\subsection{Calibrated Air Mass Setpoint Map}
\begin{figure}
    \centering
    \includegraphics[width=0.49\textwidth,keepaspectratio]{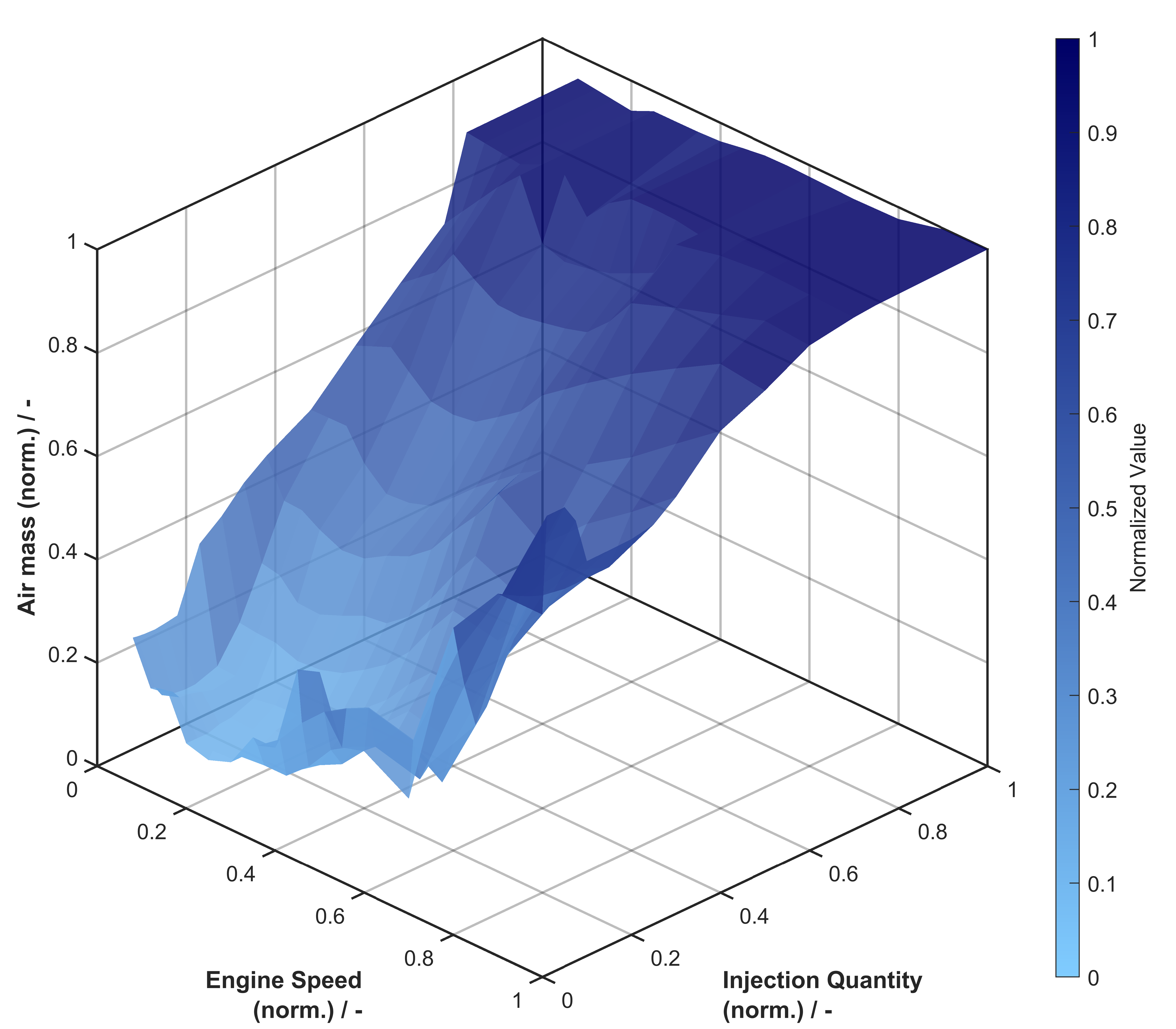}
    \caption{Resulting calibration map found by RRL-Methodology ready for deployment.}
    \label{fig:Final-Calibration}
\end{figure}

Finally, the resulting maps and their evolution throughout the individual development iterations are presented in Fig. \ref{fig:Calibration-Progress}. It becomes apparent that the agent already begins lowering the low load and low speed region in the first iteration, although only slightly at first. By the second iteration, a significant reduction in the air mass setpoint can already be observed, which leads to increased \ac{EGR} rates along with higher actuator positions for both \ac{HP} and \ac{LP} \ac{EGR}. At the same time, cumulative NOx emissions decrease by more than 20\%. Iteration steps three and four result in further reductions of the air mass setpoint to 60\% of the initial value in the part load region. However, the additional reduction of the setpoints from iteration three to four no longer yields a significant decrease in cumulative NOx emissions. Instead, a saturation effect occurs and soot emissions begin to increase. Finally, the air mass setpoint map is shown in Fig. \ref{fig:Final-Calibration}. The integrity of the original map is largely preserved, which indicates that the map could realistically be used in a vehicle. Notably, the agent already exhibits extrapolation characteristics, meaning that it alters areas that were encountered frequently during training considerably, while leaving areas that were rarely or never visited almost unchanged.

\section{Conclusion}
\label{sec:conclusion}
In this work, an explainable methodology was presented that demonstrates how an \ac{RRL} agent can be transferred into a real air mass setpoint calibration framework based on map structures. The approach consisted of multiple iterative training sessions, in which a newly trained agent was generated in each iteration, but with a deliberately constrained action space to counteract safety concerns. The final calibration exhibits comparable transient performance to the reference strategy and even outperforms the reference in terms of cumulative NOx emissions.
However, the chosen strategy results in slightly increased soot emissions, as the reward associated with boost pressure deviation and NOx reduction compensates for the negative penalty assigned to higher soot emissions. Overall, the best agent surpasses the reference calibration, achieving a reward of -570.6 compared to -574.9. This improvement is reflected in the resulting air mass setpoint calibration, which performs at a level comparable to the reference strategy. To assess this behavior, a segment of the \ac{WLTC} was evaluated, in which torque, injected fuel mass, and boost pressure were analyzed with temporal resolution. The results indicate that the \ac{RRL}‑based calibration yields performance comparable to the reference, although it differs in the selected air‑mass setpoint trajectory. Based on 295 hours of accumulated experience, the agent has learned a control strategy that differs slightly from the reference calibration.
In summary, it has been demonstrated that \ac{RRL} can be used to calibrate and optimize the air mass setpoint without requiring the deployment of neural networks on the target ECU architecture. This was validated through the iteration‑based map calibration and corresponding validation runs. As a result, new strategies can be learned and subsequently transferred into rule‑based control algorithms without sacrificing interpretability with respect to underlying physical relationships. A remaining limitation is that the reward‑function weighting factors must still be determined manually, meaning that expert knowledge remains necessary. Nevertheless, the effort required is minimal compared to a conventional manual calibration process.
The final analysis of the derived calibration map confirms the overall integrity of the map in terms of its structure and gradients, suggesting that it could even be deployed in a prototype vehicle. Looking ahead, it is conceivable that more complex maps may be calibrated using multi‑agent approaches—for instance, enabling simultaneous optimization of boost pressure control and air mass setpoint determination. The use of a \ac{HiL} system remains advantageous for such developments, as it offers an optimal balance of cost efficiency, durability, and realism.

\section*{Acknowledgements}
\label{sec:acknowledgements}
This work was carried out in parts at the \textit{Center for Mobile Propulsion
(CMP)} of the RWTH Aachen University, funded by the German Research Foundation
(DFG).
This work and the scientific research behind it have been funded by the
\emph{OPTHIK} project (grant no. EFRE-20800482) of the state of North Rhine-Westphalia on the basis of the EFRE/JTF-Program NRW. The EFRE/JTF program in North Rhine-Westphalia (NRW) promotes sustainable projects and innovations that support regional development and the transition to a climate-neutral economy.

\section*{Author Contributions}
\label{sec:author-contributions}

\textbf{Conceptualisation:} Andreas Kampmeier, Kevin Badalian, Lucas Koch;
\textbf{Methodology:} Kevin Badalian, Andreas Kampmeier;
\textbf{Software:} Kevin Badalian;
\textbf{Validation:} Andreas Kampmeier, Kevin Badalian;
\textbf{Formal analysis:} Andreas Kampemier, Kevin Badalian;
\textbf{Investigation:} Andreas Kampmeier, Kevin Badalian;
\textbf{Resources:} Jakob Andert;
\textbf{Data curation:} Andreas Kampmeier, Kevin Badalian;
\textbf{Writing -- original draft:} Andreas Kampmeier, Kevin Badalian;
\textbf{Writing -- review \& editing:} Sung-Yong Lee, Lucas Koch, Jakob Andert;
\textbf{Visualisation:} Andreas Kampmeier, Kevin Badalian;
\textbf{Supervision:} Jakob Andert;
\textbf{Project administration:} Jakob Andert;
\textbf{Funding acquisition:} Jakob Andert

% Bibliography
\printbibliography

\end{document}